\documentclass{article}

\usepackage{microtype}
\usepackage{graphicx}
\usepackage{csquotes}
\usepackage{subfigure}
\usepackage{booktabs} 

\usepackage{notation}
\usepackage{tcolorbox}

\definecolor{colorbluefull}{rgb}{0.25882352941176473, 0.5215686274509804, 0.9568627450980393}
\colorlet{colorblue}{colorbluefull!30}

\usepackage[preprint]{icml2025}

\usepackage{amsmath}
\usepackage{amssymb}
\usepackage{mathtools}
\usepackage{amsthm}

\usepackage[capitalize,noabbrev]{cleveref}

\theoremstyle{plain}

\theoremstyle{definition}
\newtheorem{definition}{Definition}

\theoremstyle{remark}

\usepackage[inline]{enumitem}

\usepackage[textsize=tiny]{todonotes}

\icmltitlerunning{On Teacher Hacking in Language Model Distillation}

\begin{document}

\twocolumn[
\icmltitle{On Teacher Hacking in Language Model Distillation}




\begin{icmlauthorlist}
\icmlauthor{Daniil Tiapkin}{x}
\icmlauthor{Daniele Calandriello}{gdm}
\icmlauthor{Johan Ferret}{gdm}
\icmlauthor{Sarah Perrin}{gdm}
\icmlauthor{Nino Vieillard}{gdm}
\icmlauthor{Alexandre Ram{\'e}}{gdm}
\icmlauthor{Mathieu Blondel}{gdm}
\end{icmlauthorlist}

\icmlaffiliation{gdm}{Google DeepMind}
\icmlaffiliation{x}{CMAP, {\'E}cole Polytechnique, Palaiseau, France; Work done during an internship at Google DeepMind.}

\icmlcorrespondingauthor{Daniil Tiapkin}{daniil.tiapkin@polytechnique.edu}

\icmlkeywords{Machine Learning, ICML}

\vskip 0.3in
]



\printAffiliationsAndNotice{\icmlEqualContribution} 

\begin{abstract}
Post-training of language models (LMs) increasingly relies on the following two stages: (i) knowledge distillation, where the LM is trained to imitate a larger teacher LM, and (ii) reinforcement learning from human feedback (RLHF), where the LM is aligned by optimizing a reward model.
In the second RLHF stage, a well-known challenge is reward hacking, where the LM over-optimizes the reward model. Such phenomenon is in line with Goodhart's law and can lead to degraded performance on the true objective.
In this paper, we investigate whether a similar phenomenon, that we call teacher hacking, can occur during knowledge distillation. This could arise because the teacher LM is itself an imperfect approximation of the true distribution. To study this, we propose a controlled experimental setup involving: (i) an oracle LM representing the ground-truth distribution, (ii) a teacher LM distilled from the oracle, and (iii) a student LM distilled from the teacher.
Our experiments reveal the following insights. When using a fixed offline dataset for distillation, teacher hacking occurs; moreover, we can detect it by observing when the optimization process deviates from polynomial convergence laws. In contrast, employing online data generation techniques effectively mitigates teacher hacking. More precisely, we identify data diversity as the key factor in preventing hacking.
Overall, our findings provide a deeper understanding of the benefits and limitations of distillation for building robust and efficient LMs.
\end{abstract}
\newpage
\section{Introduction}\label{sec:intro}

\begin{figure}
    \centering
    \includegraphics[width=0.95\linewidth]{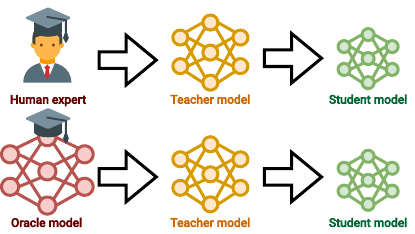}
    \caption{\textbf{Overview of our controlled experimental setup.} Usually, the teacher model is trained on expert data before being distilled into the student LM. In the controlled setup of this paper, the teacher is  itself distilled from an additional oracle model. This oracle model allows us to measure the quality of the distillation process into the student, and to reveal \enquote{teacher hacking}.}
    \label{fig:oracle_pipeline}
\end{figure}

\textbf{Distillation for post-training LMs.}
Language models (LMs) have achieved remarkable success across a wide range of natural language processing tasks, such as translation, summarization, and reasoning.
Notably, large LMs demonstrate impressive generalization capabilities, but their high computational cost poses a significant challenge, particularly when deployed on resource-constrained devices. Efficiency considerations motivate the training of smaller LMs, that would ideally provide similar performance at a fraction of the computational cost. To this end, the most popular approach is knowledge distillation (KD) \cite{hinton2015distilling}, in which a smaller student LM is trained to imitate the larger teacher LM.
Distillation is increasingly studied \cite{agarwal2024onpolicy,gu2024minillm,kim2024promptkd} and used, notably for the post-training pipelines of LMs (as demonstrated by Zephyr \cite{tunstall2023zephyr}, Gemma-2 \cite{team2024gemma}, and DeepSeek-V3 \cite{liu2024deepseek}), just before the final reinforcement learning from human feedback (RLHF) \cite{stiennon2020learning,ouyang2022training,bai2022training} phase.

\textbf{Teacher as an imperfect proxy.}
However, a key understudied limitation of KD is that the teacher model does not represent the ground-truth distribution but instead acts as an imperfect proxy for it \citep{menon2021statistical,zhang2023not}. This viewpoint draws parallels to a well-studied phenomenon in RLHF known as reward hacking \cite{amodei2016concrete,pan2022effects,gao2023scaling}. Reward hacking is a manifestation of Goodhart's law arising when LMs over-optimize the reward model, trained to represent human preference, thus also an imperfect proxy for the true task objective. The consequences of reward hacking in RLHF can be significant, leading to models misleading humans and producing unsafe behaviors \cite{hendrycks2021unsolved,wen2024language}.

\textbf{Teacher hacking.}
Inspired by the analogy with reward hacking in RLHF, we define teacher hacking as a possible phenomenon during distillation in which the student LM learns to imitate the teacher, not by better approximating the true data distribution, but by exploiting imperfections in the teacher.
This raises natural questions: \textit{(1) Does teacher hacking occur in practice?} and if so, \textit{(2) when does it appear?}, and \textit{(3) what strategies can mitigate it?}

\textbf{Controlled experimental setup.}
To answer these questions, we propose a controlled experimental setup. Specifically, we introduce an oracle model that represents the ground-truth; see \Cref{fig:oracle_pipeline} for an illustration. In this setup, the distance between the student and oracle LMs serves as a \enquote{golden metric}. Conversely, the distance between student and teacher LMs, optimized during the fine-tuning of the student, defines a \enquote{proxy metric}. 
Our setup is semi-synthetic, in the sense that prompts are sampled from real datasets but responses are sampled from the oracle LM to learn the teacher, and from the teacher LM to learn the student.

We summarize our main contributions as follows.
\begin{itemize}[topsep=0pt,itemsep=3pt,parsep=3pt,leftmargin=15pt]
\item
We introduce and formally define the phenomenon of teacher hacking in LM distillation, providing a novel controlled experimental framework to systematically measure its importance.
\item
We show that teacher hacking occurs when distillation is performed on a fixed offline dataset. Notably, we empirically found that the proxy metric initially follows a polynomial convergence law, and that teacher hacking emerges when the optimization process deviates from this law.
\item
We analyze strategies to mitigate teacher hacking.
Notably, we show that teacher hacking is absent when using online data generation methods, sampling directly from the teacher or the student during distillation. We relate this success to increased diversity in the dataset. We then propose cheaper practical alternative strategies to mitigate teacher hacking, such as increasing the diversity across prompts or using a larger amount of offline teacher-generated data.
\end{itemize}

\section{Preliminaries}\label{sec:preliminaries}

Let $\cX$ and $\cY$ denote the spaces of all possible prompts and responses, assumed to be sentences in the vocabulary $\Sigma$. For two sets \(A\) and \(B\), \(\simplex(A)\) denotes the space of probability measures over \(A\), and \(\simplex(A|B)\) represents the space of conditional probability measures over \(A\) given \(B\). An auto-regressive language model \(\pi \in \LM_{\Sigma}(\cX)\) is defined as the conditional probability of the next token or the end-of-sequence token given a prompt \(x \in \mathcal{X}\) and a partial generation \(y \in \cY\), expressed as \(\pi(\omega | x, y)\), where $\omega \in \Sigma$. In practice, the probability $\pi(\omega | x,y)$ is defined using a softmax with temperature $\pi(\omega | x,y) \propto \exp( \frac{1}{\tau}z(\omega| x, y))$, where $z(\cdot | x,y)$ are the logits output by the neural network and $\tau$ is a temperature parameter.

For any language model \(\pi\), the induced distribution over responses is given by
$
    p_{\pi}(y|x) \triangleq \prod_{i=1}^{|y|} \pi(y_i | x, y_{:i})\,,
$
where $|y|$ denotes the length of the response \(y\), i.e., the number of tokens (non-empty characters) in \(y\), and $y_{:i} \triangleq (y_1, y_2, \ldots, y_{i-1})$.

\paragraph{Distances between language model distributions.} Let $d \in \simplex(\cX)$ be a distribution over prompts induced by the particular task dataset. To measure the convergence of one distribution, induced by a language model $\pi$, to a distribution induced by a language model $\pi'$, we use the (expected) forward and reverse KL between the corresponding conditional measures $p_{\pi}$ and $p_{\pi'}$, defined as $\KL_{\seq}(\pi,  \pi') \triangleq \E_{x \sim d(\cdot)}[\KL(p_{\pi}(\cdot|x) , p_{\pi'}(\cdot|x))]$. By the properties of the KL divergence, the sequence-level divergence can be estimated very efficiently using token-level KL divergences. Additionally, we use a sequence-level Jensen-Shannon divergence $\JS_{\seq}(\pi, \pi')$. We provide more details in~\Cref{app:misc}.


\paragraph{Supervised fine-tuning.} Let $\rho \in \simplex(\cY | \cX)$ be a conditional response distribution that encodes the ground-truth response distribution. Solving downstream tasks such as summarization, translation, reasoning, or instruction following is fundamentally equivalent to approximating $\rho$. Therefore, the ultimate goal of any post-training pipeline is to approximate $\rho$ in order to address these tasks effectively. 

One of the common approaches to this problem is supervised fine-tuning (SFT). Let us assume that we have a dataset of pairs $(x,y)$ for $x \sim d(\cdot)$ and $y \sim \rho(\cdot | x)$. Then, to find a language model $\pi$ such that its conditional distribution $p_{\pi}$ approximates $\rho$, it is common to use a simple log-loss
\begin{equation}\label{eq:sft_loss}
    \cL_{\mathrm{SFT}}(\pi) \triangleq \E_{x \sim d(\cdot), y \sim \rho(\cdot | x)}\left[ - \log p_{\pi}(y|x) \right]\,.
\end{equation}
This loss is equal, up to a constant factor, to an expected sequence-level forward KL divergence between $\rho$ and $p_{\pi}$: $\E_{x \sim d(\cdot)}[\KL(\rho, p_{\pi})]$.


\paragraph{Language model distillation.}

We suppose that we have access to a teacher language model,
denoted $\pi_{\teacher} \in \LM_\Sigma(\cX)$, such that it approximates the ground-truth distribution $\rho$, that is, $p_{\teacher} = p_{\pi_{\teacher}} \approx \rho$. The goal of language model distillation is to train a student language model, denoted $\pi_{\student}$, so as to approximate the teacher model, that is, $p_{\student} = p_{\pi_{\student}} \approx p_{\teacher}$. We emphasize that the teacher-induced distribution $p_{\teacher}$ is not equal to a ground-truth $\rho$ but only approximates it. 

We usually distinguish between \textit{hard} and \textit{soft} distillation. In hard distillation, the student is only trained from the teacher's predicted next tokens, i.e., the student only sees the most likely tokens according to the teacher. In soft distillation, the student is trained from the teacher's predicted next token distributions, giving much more information to the student. In this work, we focus on soft distillation, and the loss function for this procedure takes the form
\begin{equation}\label{eq:distillation_loss}
    \cL_{\mathrm{KD}}(\pi_{\student}) \triangleq \E\left[ \frac{1}{|y|} \sum_{i=1}^{|y|} \ell_{\mathrm{token}}(\pi_{\student}(\cdot | x, y_{:i}), \pi_{\teacher}(\cdot | x, y_{:i})) \right]\,,
\end{equation}
where $x \sim d(\cdot)$ and $y \sim \nu(\cdot| x)$, \(\nu \in \simplex(\mathcal{Y} | \mathcal{X})\) is a \textit{data source}, and \(\ell_{\mathrm{token}}\) is a token-level loss function between two distributions over the vocabulary.

The token-level loss and the data source should satisfy two assumptions: (i) it is non-negative and satisfies the property $\ell_{\mathrm{token}}(p,q) = 0$ if and only if $p = q$ for any two distributions $p,q \in \simplex(\Sigma)$, and (ii) the support of $\nu(\cdot | x)$ includes the support of teacher-induced conditions measure $p_{\teacher}(\cdot | x)$ for almost all $x$, i.e. $d(x) > 0$.  Given these two assumptions, it is easy to show that a language model $\pi$ minimizes the loss $\cL_{\mathrm{KD}}(\pi_{\student}) = 0$ if and only if $\pi_{\student} = \pi_{\teacher}$. 

In particular, considering $\nu(\cdot|x)$ induced by an offline dataset generated by a teacher and using $\ell_{\mathrm{token}}(p,q) = \KL(p,q)$, we achieve the same expected loss as in the case of SFT. This approach corresponds to the works of \cite{hinton2015distilling, sanh2019distilbert}. However, we could consider different token-level losses, such as reverse KL divergence \cite{gu2024minillm}, generalized Jensen-Shannon divergence \cite{agarwal2024onpolicy}, or skewed KL divergence \cite{ko2024distillm}. The data source can be induced by sampling online from the teacher model \cite{kim2016sequence}, sampling online from the student model \cite{gu2024minillm}, or combining offline and online data \cite{lin2020autoregressive, agarwal2024onpolicy, ko2024distillm}.

\paragraph{Offline vs. online data sources.}

In this paper, we distinguish between two different types of data sources: offline and online.
Offline data sources are based on a fixed dataset, denoted as $\cD_{\mathrm{offline}} = \{(x_i, y_i)\}_{i=1}^M$, where $x_i$ is sampled from $\cD_{\mathrm{prompt}}$, and $y_i \sim p_{\teacher}(\cdot | x_i)$ are responses generated by the teacher model. Importantly, the dataset $\cD_{\mathrm{offline}}$ does not need to have a one-to-one correspondence between prompts and responses. Instead, it may include multiple responses per prompt, and the number of prompts is not necessarily equal to the total number of available prompts $N$. Each training batch is sampled directly from this fixed dataset.

Online data sources, by contrast, involve generating responses dynamically through an online sampling procedure. For each training batch of prompts $\{x_j\}_{j=1}^B$ of size $B$, sampled from $\cD_{\mathrm{prompt}}$, a corresponding model (either the teacher or the student) generates a batch of responses $\{y_j\}_{j=1}^B$. While sampling from the student model is often referred to as on-policy generation in the literature \cite{agarwal2024onpolicy}, we use the term online student to emphasize its parallel to the online teacher data source. 

The distinction between offline and online data is particularly evident during subsequent epochs: for offline data sources, responses remain fixed across epochs, while for online data sources, new responses are generated for each epoch, independently drawn from the same distribution.

\paragraph{Teacher hacking.} 
As we already mentioned, the main goal of the post-training pipeline is to approximate the ground-truth distribution $\rho$ by the student model. This goal could be achieved by SFT given access to sufficiently many samples from $\rho$. However, recent works have shown that distilling from a teacher model can actually work better since the whole next-token distribution contains much richer information than only sampled tokens. An understudied problem is that the teacher model is only an imperfect proxy for $\rho$. This may lead to a problematic situation where the student LM learns to imitate the teacher, not by better approximating the true data distribution but by exploiting imperfections in the teacher.
We call this phenomenon teacher hacking, and give a more formal definition below.
\begin{tcolorbox}[colback=colorblue,
    colframe=black,
    arc=4pt,
    boxsep=0.3pt,
]%
\begin{definition}[Teacher hacking]\label{def:teacher_hacking}
    Let $\{p_{\student}^{(k)}\}_{k=1}^\infty$ be a sequence of conditional response distributions induced during the training of a student model, $p_{\teacher}$ the distribution induced by the teacher model, and $\rho$ the target human expert conditional distribution. We say that $\{p_{\student}^{(k)}\}_{k=1}^\infty$ exhibits the teacher hacking phenomenon with respect to a distance measure $\dist \colon \simplex(\cY | \cX) \times \simplex(\cY | \cX) \to \mathbb{R}_{+}$ if, as $k \to +\infty$, $\dist(p_{\student}^{(k)}, p_{\teacher})$ decreases while $\dist(p_{\student}^{(k)}, \rho)$ increases.
\end{definition}
\end{tcolorbox}

A simple example where this would occur is when the student model is initially closer to the target distribution $\rho$ than the teacher model. However, in more realistic scenarios that are closer to real-world applications, the teacher model is larger and provides a better approximation of $\rho$ than the student model.

\paragraph{Overfitting vs. teacher hacking.} 

We would like to clarify the difference between two related phenomena: classical overfitting and the newly-introduced teacher hacking. In the case of overfitting, the model continues to minimize the loss on the training set but fails to do so on the held-out validation set, which it never observes. In contrast, teacher hacking occurs when the model successfully achieves its objective from the teacher's perspective, even on the validation dataset, but fails to improve in terms of approximating the ground-truth behavior. The ultimate outcome of both phenomena is similar: the model fails to generalize to the ground truth, but the reasons differ.
\section{Methodology}\label{sec:methodology}
\begin{figure}
    \centering
    \includegraphics[width=0.95\linewidth]{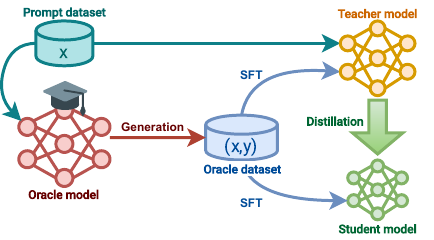}
    \caption{\textbf{Overview of the training pipeline.} Two stages: (1) prompts $x$ from a task-specific real dataset are used by the oracle model to generate the oracle pairs $(x,y)$, and afterwards, this dataset is used to get initial SFT checkpoints for both teacher and student model; (2) prompts from the same distribution are used to perform knowledge distillation, where the teacher model serves as a proxy to train the student model.}
    \label{fig:training_pipeline}
\end{figure}

To analyze the effect of teacher hacking, we require a method to estimate the distance between the student model and the ground-truth distribution. For this purpose, we introduce the \textit{oracle model}, denoted as $\mu \in \LM_\Sigma(\cX)$, which is assumed to induce the target distribution $\rho$, i.e., $p_{\mu} = \rho$.

\paragraph{Golden and proxy metrics.} As outlined in \Cref{def:teacher_hacking}, evaluating the teacher hacking phenomenon requires computing two sets of metrics. 

Golden metrics are computed using the oracle model and reflect the performance with respect to the true objective. Specifically, we use three types of divergences: the forward KL divergence $\KL_{\seq}(\mu, \pi_{\student})$, the reverse KL divergence $\KL_{\seq}(\pi_{\student}, \mu)$, and a Jensen-Shannon-like distance $\JS_{\seq}(\pi_{\student}, \mu)$, which is closely related to the Jensen-Shannon divergence between the conditional distributions over the response space. These metrics are estimated using a held-out validation set of prompts, with sampling performed from the respective models.

Proxy metrics, in contrast, do not rely on the oracle model and instead measure the alignment between the student and teacher models. These metrics use the same types of distances: the forward KL divergence $\KL_{\seq}(\pi_{\teacher}, \pi_{\student})$, the reverse KL divergence $\KL_{\seq}(\pi_{\student}, \pi_{\teacher})$, and the Jensen-Shannon divergence $\JS_{\seq}(\pi_{\student}, \pi_{\teacher})$. Similar to the golden metrics, proxy metrics are estimated using the validation set of prompts and sampling from the models involved.

\paragraph{Training.} The training procedure for our experiments consists of two stages.

In the first stage, supervised fine-tuning is performed on both the teacher and student models using a small oracle-generated dataset $\cD_{\mathrm{oracle}} = \{(x_i, y_i)\}_{i=1}^{N_{\mathrm{oracle}}}$. The prompts $x_i$ are sampled from the task distribution $d(\cdot)$, and the responses $y_i \sim \rho(\cdot | x_i)$ are generated using the oracle model. 
Our setup is semi-synthetic, in the sense that prompts are sampled from real datasets but responses (seen as labels) are sampled from the oracle LM to learn the teacher, and from the teacher LM to learn the student.
This stage is the only place where direct information from the oracle model is propagated to the teacher and student models. The fine-tuning process optimizes a usual SFT loss \eqref{eq:sft_loss} that in expectation equals to a sequence-level distance $\KL_{\seq}(\mu, \pi)$. The best checkpoint is selected based on the estimate of this quantity over the validation set.

In the second stage, distillation is conducted from the teacher to the student model by optimizing the soft-distillation loss \eqref{eq:distillation_loss}. The distillation process uses a training dataset of unlabeled prompts $\cD_{\mathrm{prompt}} = \{x_i\}_{i=1}^N$, where $N \gg N_{\mathrm{oracle}}$ and different data sources that define a distribution of $y_i \sim \nu(\cdot | x)$ in the loss. The final pipeline is summarized in \Cref{fig:training_pipeline}.

\begin{figure}
    \centering
    \includegraphics[width=0.95\linewidth]{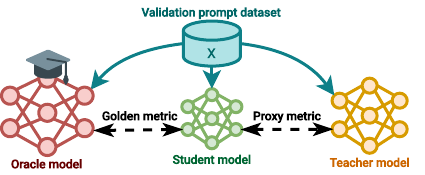}
    \caption{\textbf{Overview of the evaluation pipeline.} We use the validation prompt dataset to measure the golden metric (the distance between the oracle and the student models) and the proxy metric (the distance between the teacher and the student models).}
    \label{fig:evaluation_pipeline}
\end{figure}

\paragraph{Evaluation.}
To investigate the teacher hacking phenomenon, we analyze two key types of curves: (1) the dependence of the training loss, proxy metrics, and golden metrics on the number of epochs completed, and (2) the proxy-golden curve, which illustrates the relationship between the golden metric (only accessible in our controlled experimental setup) and the proxy metric. Both proxy and golden metrics are computed using a held-out validation set of prompts. The final pipeline is summarized in \Cref{fig:evaluation_pipeline}.

The epoch-dependence plots provide insights into scaling law phenomena \cite{kaplan2020scaling} and help to understand overall training dynamics. Proxy-golden curves are crucial for visually assessing the presence of teacher hacking: a U-shaped curve serves as a clear indicator. Indeed, we expect the proxy metric to be reduced during training, and if the golden metric first decreases and then increases, it directly shows teacher hacking. These proxy-golden plots can be compared to plots from \citet{gao2023scaling}, with one essential difference: our approach measures optimization progress as the distance to the teacher model rather than the distance from an initial reference policy.

\section{Experimental results}\label{sec:results}



\paragraph{Oracle, teacher, and student models.} Our experiments use a family of encoder-decoder language models based on T5 \cite{raffel2020exploring,roberts2022t5x}. The oracle model is the Flan-T5-XL \cite{chung2024scaling}, a 3B-parameter model fine-tuned on the Flan dataset \cite{wei2021finetuned,longpre2023flan} for instruction-based tasks. For the teacher and student models, we use pretrained checkpoints of T5-1.1 in three configurations: small (77M parameters), base (250M parameters), and large (800M parameters). We always use temperature sampling with a temperature parameter $\tau = 1$ for generations from any model.

\paragraph{Datasets.} Our experiments use three datasets for training and evaluation: the XSum summarization dataset \cite{narayan2018dont}, the WMT-14 en-de translation dataset \cite{bojar2014findings}, and the instruction-following dataset Natural Instructions \cite{naturalinstructions,supernaturalinstructions}. In alignment with our experimental setup, we use only the prompt data from these datasets, supplemented with task-specific instructions as needed for each task.

For the first stage of the training pipeline, where the oracle dataset is build and used for SFT, we use $N_{\mathrm{oracle}} = 25\,000$, $50\,000$, and $100\,000$ prompts from the XSum, WMT-14 en-de, and Natural Instructions datasets, respectively. For the second stage, which involves the knowledge distillation procedure, we use $N = 200\,000$, $450\,000$, and $500\,000$ prompts from these datasets. A single epoch is defined as one complete pass through all $N$ examples, corresponding to $\lceil N / B \rceil$ training steps, where $B$ denotes the batch size. For XSum and WMT-14 en-de, we use batch size $B=32$; for Natural Instructions, we use batch size $B=64$.

\subsection{Does teacher hacking appear?}\label{sec:online_vs_offline}

\begin{figure}[ht]
    \centering
    \includegraphics[width=0.99\linewidth]{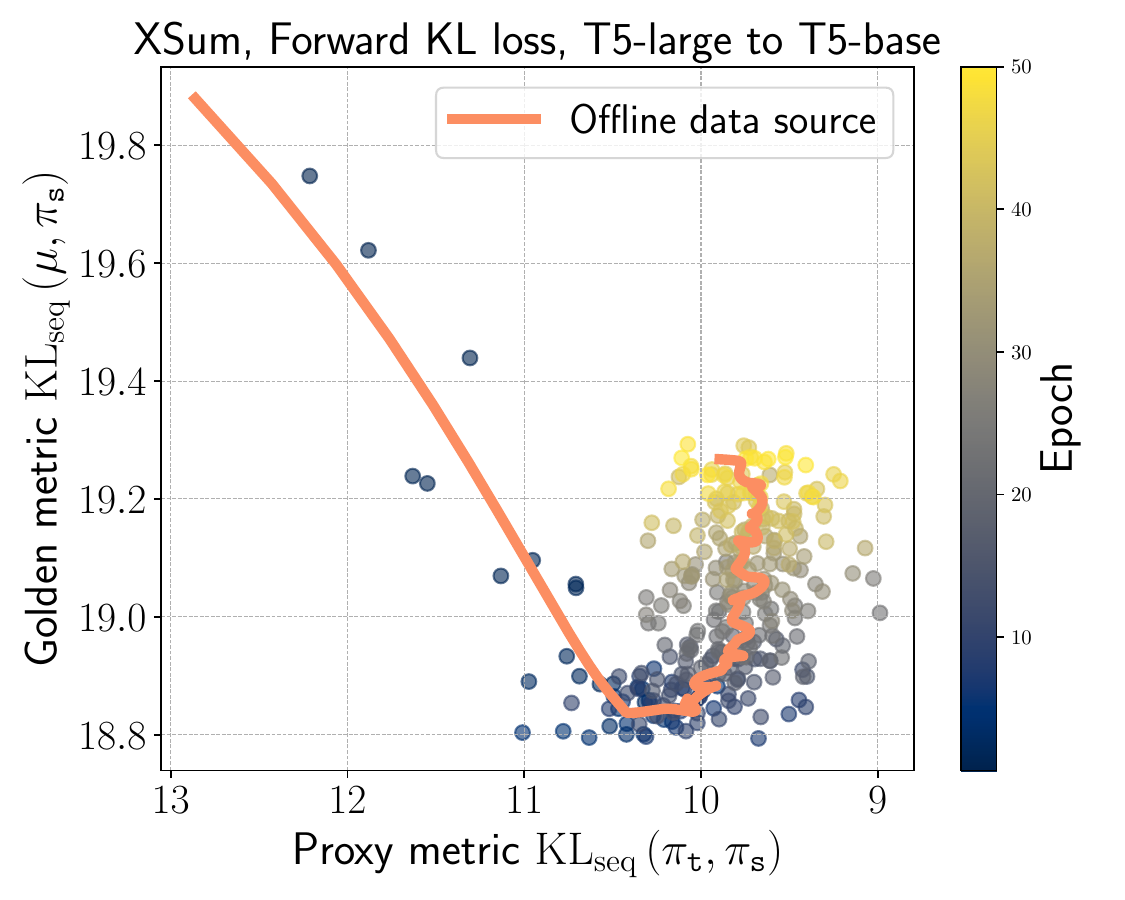}

    \caption{\textbf{Proxy-Golden plot (offline data source).}
    We distill a T5-large teacher into a T5-base student on the XSUM dataset. The token-level training loss is the forward KL, the proxy metric is the distance to the teacher distribution and the golden metric is the distance to the ground-truth (oracle) distribution (available thanks to our semi-synthetic controlled experimental setup).
    In this plot, the $x$-axis (proxy metric) indicates optimization progress, and the $y$-axis shows the ground-truth performance (golden metric): lower is better. Teacher hacking occurs in the case of offline data source: the orange curve has a U-type shape, indicating that during optimization, the orange metric starts increasing, whereas the proxy metric continues to decrease.
    }
    \label{fig:proxy_golden_exp_large_to_base_offline_vs_online_fwd_kl}
\end{figure}

\begin{figure*}[ht]
    \centering
    \includegraphics[width=0.99\linewidth]{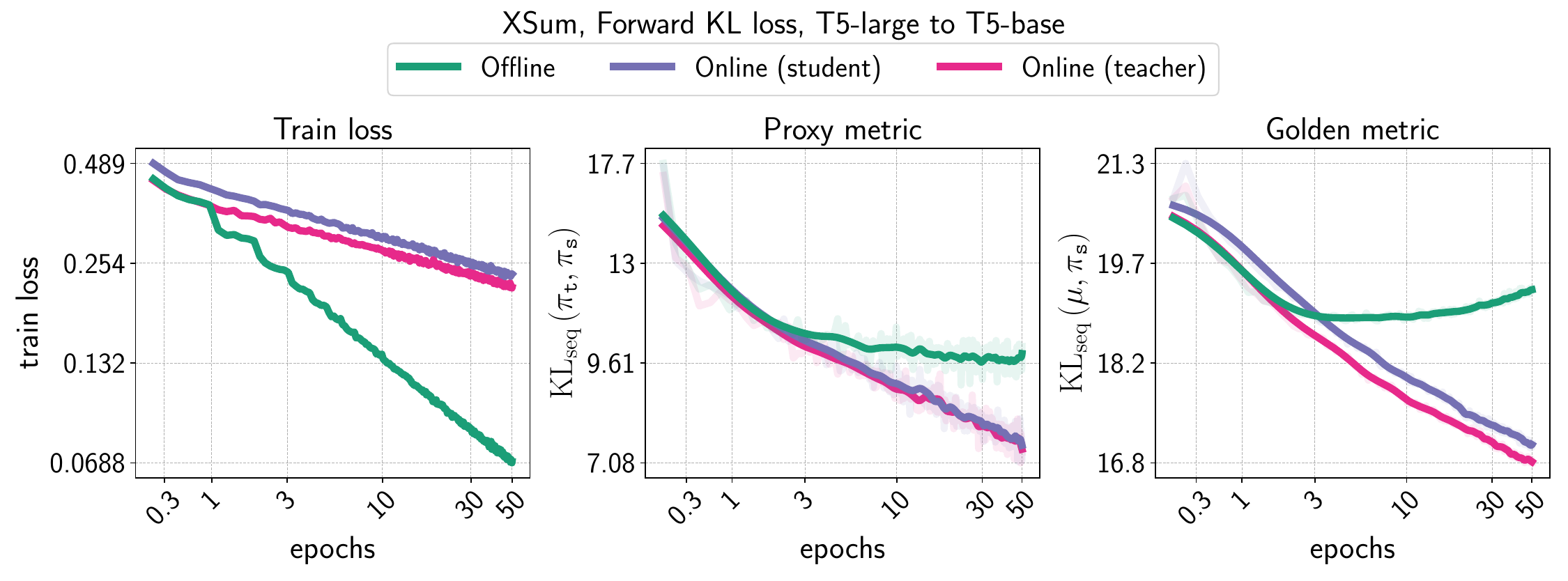}
    \caption{\textbf{Impact of using offline vs. online data sources.} 
     When using a fixed offline dataset, though the proxy metric continues to decrease, this is not visible in the golden metric, which continues to increase, a phenomenon we call teacher hacking. However, when using online response sampling, both from the teacher model or from the student model, this phenomenon does not occur.
    }
    \label{fig:main_exp_xsum_fwd_kl_base_large}
\end{figure*}

We begin by investigating whether teacher hacking appears. 

\paragraph{Setup.}
\begin{figure*}[ht]
    \centering
    \includegraphics[width=0.99\linewidth]{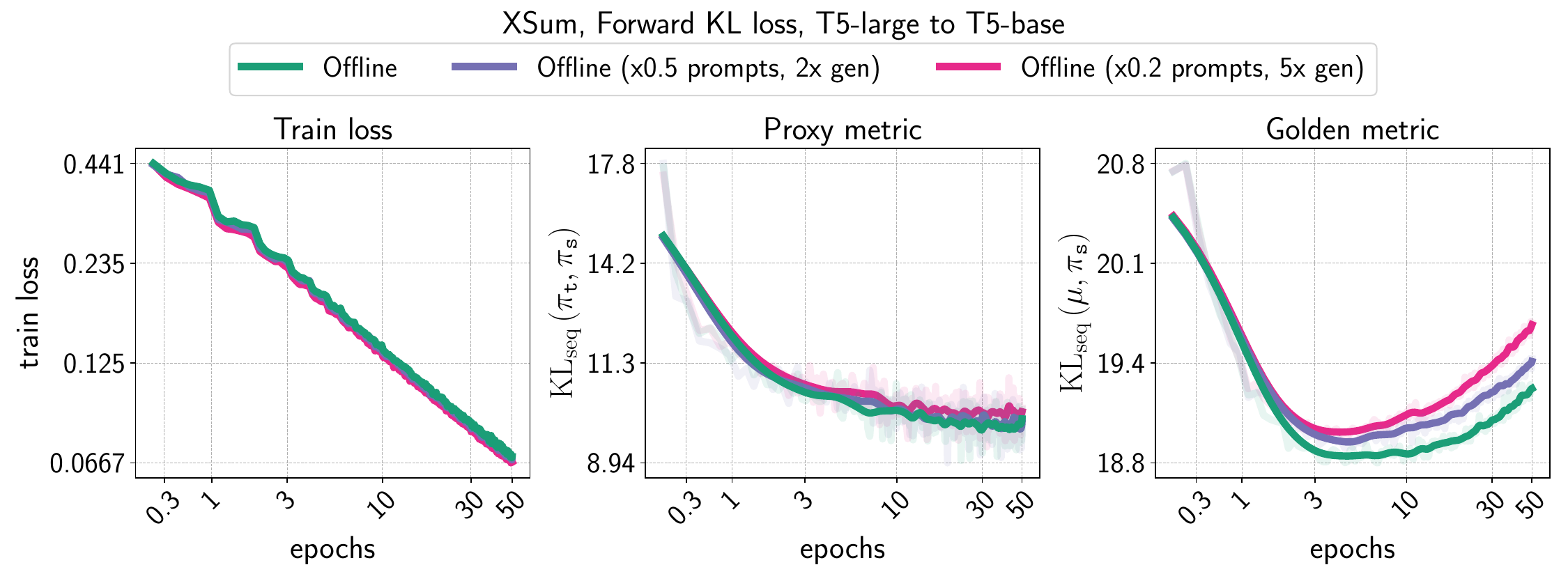}
    \caption{\textbf{Impact of diversity of offline data sources.} 
    We regulate the diversity of the dataset by decreasing the number of prompts in $2/5$ times and providing $2/5$-times more generations for each existing prompt, while preserving the size of the dataset.
    Whereas the dynamics of the train loss and proxy metric are almost the same, the effect of teacher hacking becomes more evident with a less diverse dataset.    
    }
    \label{fig:exp_large_to_base_diversity_fwd_kl}

\end{figure*}

For the first experiment, we use only \textit{offline data sources}: responses are pre-generated as $y_i \sim p_{\teacher}(x_i)$ for all $x_i \in \cD_{\mathrm{prompt}}$, and the dataset remains fixed throughout training. The learning rate for optimization is selected via a grid search over $\{ 10^{-4}, 3 \times 10^{-4}, 10^{-3} \}$.

The distillation procedure starts from the SFT checkpoints of the teacher and student models. Training is carried out over 50 epochs to analyze long-term convergence behavior.

\paragraph{Results.}
The results of distilling the T5-large teacher model into the T5-base student on the XSum dataset, using forward KL loss, along with the corresponding golden and proxy metrics, are shown in \Cref{fig:proxy_golden_exp_large_to_base_offline_vs_online_fwd_kl}.


In this plot, the $x$-axis represents the optimization progress in terms of the distance to the teacher model (from left to right), while the $y$-axis shows the golden metric. The scatter plot shows the exact values of proxy and golden metrics, where the color demonstrates at which epoch this measurement was performed. The curve itself shows a relationship between smoothed values of the proxy and golden metric, where smoothing is performed by Gaussian smoothing.

For offline data source, the plot exhibits a U-shaped curve. This behavior indicates teacher hacking: as optimization progresses, the ground-truth performance (golden metric) initially improves but eventually deteriorates. Overall, the conclusion of this experiments is following.

\begin{tcolorbox}[colback=colorblue,
    colframe=black,
    arc=4pt,
    boxsep=0.3pt,
]%
\textbf{Observation 1.} Teacher hacking exists and emerges after extended training on a fixed offline dataset.
\end{tcolorbox}

\subsection{When does teacher hacking appear?}
In this subsection, we investigate the conditions under which teacher hacking occurs.

\paragraph{Setup.}

For this experiment, we evaluate three distinct data sources: (1) \textit{Offline data source};
(2) \textit{Online teacher data source}: for each batch of prompts sampled from $\cD_{\mathrm{prompt}}$, a new response $y_i \sim p_{\teacher}(x_i)$ is dynamically generated by the teacher model;
(3) \textit{Online student data source}: responses are generated on-the-fly as $y_i \sim p_{\student}(x_i)$ using the current student model.

As in the previous experiment, we use the forward KL divergence as a token-level loss. We refer to \Cref{app:add_plots} for different token-level loss functions, such as reverse KL divergence and Jensen-Shannon divergence.

We analyze the dynamics of proxy and golden metrics for each data source. For the proxy and golden metrics, we apply Gaussian smoothing to smooth the noisy behavior of the curves. We would also like to emphasize that the difference in scaling between the training loss and proxy/golden metrics is due to averaging over sentence length, which is used in training loss but not in proxy/golden metrics. The training loss indicates the token-level training convergence, whereas the proxy/golden metrics show the sentence-level validation convergence.

\paragraph{Results.} 

The comparison results between offline and online data sources are shown in \Cref{fig:main_exp_xsum_fwd_kl_base_large}. Additional comparisons for other combinations of datasets, student/teacher model sizes, and loss functions are provided in \Cref{app:add_plots}.

We analyze the epoch dependence of the training loss, proxy metric, and golden metric. We present the dependencies on a log-log scale to track possible polynomial convergence laws: the polynomial dependence of the metric on the training time. Overall, we make the following observations.

\textbf{(i)} In the offline data scenario, we validate the presence of teacher hacking: the proxy metric decreases while the golden metric increases after a certain point. This phenomenon does not occur with online data sources.

\textbf{(ii)} We notice that the behavior of all curves for online and offline data sources is different. Overall, the training loss for the offline data sources decreases faster since the training loss is optimized multiple times over the same data, but the performance on the proxy/golden metrics is worse overall.

\textbf{(iii)} The proxy metric for online data sources follows a linear trend on the log-log scale, indicating a polynomial convergence law. In contrast, teacher hacking in offline data coincides with deviations in the proxy metric compared to online data. It gives a mechanism to \textit{detect} teacher hacking using only the proxy metric, that is measurable even in real scenarios (not only in our controlled experimental setup).

These results highlight that teacher hacking can harm ground-truth performance, particularly when training involves multiple epochs on the same dataset. However, we would like to emphasize that this issue is not present when training is limited to a small number of epochs (e.g., 1–3), as the golden metric remains stable in these cases. We summarize the conclusions of these first experiments as follows.

\begin{tcolorbox}[colback=colorblue,
    colframe=black,
    arc=4pt,
    boxsep=0.3pt,
]%
\textbf{Observation 2.} Employing online data generation or limiting training to a few epochs effectively prevents teacher hacking.
\end{tcolorbox}

\subsection{How to mitigate teacher hacking?}\label{sec:diversity}

\begin{figure*}[ht]
    \centering
    \includegraphics[width=0.95\linewidth]{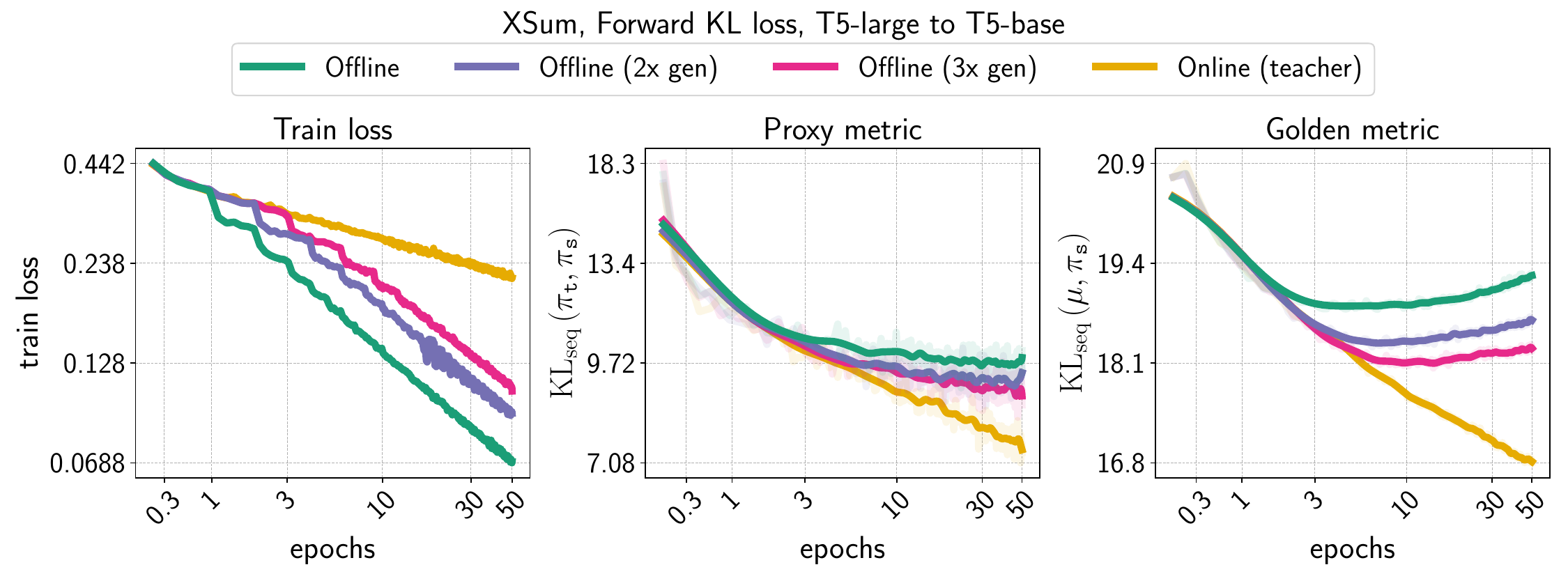}
    \caption{\textbf{Impact of generation budget for offline data sources.} 
    As the number of generations per prompt increases, both proxy and golden metrics improve, suggesting that the effect of teacher hacking is decreasing.
    }
    \label{fig:exp_large_to_base_budget_fwd_kl}
\end{figure*}

In the next experiment, we evaluate different methods for modifying the diversity and amount of offline data, in order to investigate how the properties of the offline data affect the teacher hacking phenomenon.

\paragraph{Setup.}

For this experiment, we evaluate different approaches to constructing an offline data source. We define an \textit{ordinary} offline data source as one that uses all available prompts and a single generation for each prompt: $\cD_{\mathrm{offline}} = \{ (x_i, y_i) \mid y_i \sim p_{\teacher}(\cdot \mid x_i) \}_{i=1}^N$, where $\cD_{\mathrm{prompt}} = \{ x_i \}_{i=1}^N$.

Our first objective is to study how the diversity of the offline dataset impacts the teacher hacking phenomenon under a fixed dataset generation budget. Let $k \in \mathbb{N}$ be a natural number. To construct a dataset with reduced diversity, we sub-sample $\lceil N / k \rceil$ prompts from $\cD_{\mathrm{prompt}}$ and generate $k$ responses for each sampled prompt using the teacher model. The resulting dataset maintains the same generation budget of $N$ total responses but exhibits reduced diversity because the $k$ generations for the same prompt $x$ are closer to each other compared to $k$ generations for $k$ different prompts. In our experiments, we apply this technique for $k = 2$ and $k = 5$.

Second, we investigate how increasing the generation budget influences teacher hacking. For a fixed integer $m \in \mathbb{N}$, we generate $m$ responses for each prompt in $\cD_{\mathrm{prompt}}$, resulting in a dataset of size $m \times N$. Despite the larger dataset size, we define epochs as passing through $N$ data points to ensure comparability across experiments. This setup enables us to interpolate between using an offline data source and an online teacher data source. We use values $m=2$ and $m=3$ for our experiments. The rest of the experimental setup follows the description in \Cref{sec:online_vs_offline}.

\paragraph{Diversity of offline data sources.}

We begin by examining the impact of dataset diversity on training dynamics and the previously observed teacher hacking phenomenon. The results are shown in \Cref{fig:exp_large_to_base_diversity_fwd_kl}. Notably, while the training loss and proxy metric dynamics remain nearly the same, the golden metric behaves differently: lower dataset diversity leads to worse golden metric performance, making the teacher hacking effect more evident.

\paragraph{Generation budget.}
Next, we examine the effect of a larger generation budget on training dynamics, as shown in \Cref{fig:exp_large_to_base_budget_fwd_kl}. Increasing the number of generations per prompt uniformly improves both the proxy and golden metrics over time and, oppositely, alters the training loss dynamics. Overall, this suggests an interpolation between the behavior of an ordinary offline and online teacher data sources.

\paragraph{Discussion.}

The results suggest two practical strategies to mitigate the effects of teacher hacking:

\begin{tcolorbox}[colback=colorblue,
    colframe=black,
    arc=4pt,
    boxsep=0.3pt,
]%
\textbf{Observation 3.} \textit{Prioritize Prompt Diversity.} When the generation budget for the distillation dataset is fixed, focusing on increasing the diversity of prompts can help reduce the impact of teacher hacking.
\end{tcolorbox}

\begin{tcolorbox}[colback=colorblue,
    colframe=black,
    arc=4pt,
    boxsep=0.3pt,
]%
\textbf{Observation 4.} \textit{Expand the Dataset with Multiple Completions.} If the prompt dataset is fixed, increasing the generation budget by generating multiple completions per prompt also helps diminish the effects of teacher hacking.
\end{tcolorbox}

\section{Related work}\label{sec:related_work}

Goodhart's law states: \enquote{When a measure becomes a target, it ceases to be a good measure} \cite{strathern1997improving}. In particular, it manifests itself as reward hacking in RLHF \cite{amodei2016concrete,gao2023scaling,weng2024rewardhack}. 
A line of works studied reward hacking under controlled experimental setups \cite{gao2023scaling,rafailov2024scaling}. 
Our setup closely resembles that of \citet{gao2023scaling}, where two types of reward models (RMs) are used: a golden reward model, which substitutes the ground-truth reward function, and a proxy reward model, trained on golden RM-preferred generations as ground-truth preferences. Specifically, we employ oracle and teacher models in the same roles as the golden and proxy RMs, respectively: the teacher model is trained on oracle-generated data, while the final student model is trained using the teacher's next-token distribution. 

Given possible negative consequences of reward hacking \cite{hendrycks2021unsolved,wen2024language}, another line of research attempts to mitigate its effects through better reward modeling or more robust training procedures \cite{chen2024odin,rame2024warm,liu2024rrm}.

Knowledge distillation (KD) was initially introduced as a method to compress a large teacher model into a smaller student model without much loss in performance \cite{buciluǎ2006model,hinton2015distilling}; it has then been successfully used to create smaller language models such as DistilBERT \cite{sanh2019distilbert},  Zephyr \cite{tunstall2023zephyr}, Gemma-2 \cite{team2024gemma}, Gemini-1.5 Flash \cite{team2024gemini}, and DeepSeek-V3 \cite{liu2024deepseek}. One of the actively used approaches to language model distillation is sequence-level KD \cite{kim2016sequence}. In our case, this approach corresponds to utilizing offline data sources and forward KL token-level loss (thanks to properties of the KL). Other approaches focus on matching different quantities of the teacher model by the student model, such as hidden states \cite{jiao2019tinybert} or attention scores \cite{wang2020minilm}.


Most of the previously-mentioned works implicitly or explicitly assume that the replication of the teacher model represents the final goal of the distillation process, contrary to the approach mentioned by \citet{menon2021statistical}, where the teacher is only an imperfect approximation of the true data distribution. Based on this perspective, \citet{zhang2023not} developed a perturbed loss that can be seen as training from a \enquote{proxy teacher}. However, they do not further investigate the consequences of over-optimizing the objective.

\section{Conclusion}\label{sec:conclusion}
In this paper, we introduce and examine the phenomenon of teacher hacking in language model distillation by designing a semi-synthetic controlled experimental setup. This allows us to measure its effects, and validate experimentally its presence when using a fixed offline dataset for the distillation procedure. 

Fortunately, as a practical outcome of our study, we were able to identify several strategies to mitigate teacher hacking: (1) utilize online generations during the distillation process, (2) when the generation budget is fixed, prioritize increasing the diversity of the prompt dataset, and (3) if the prompt dataset is fixed and online generations are not feasible, generate multiple offline completions per prompt ahead of time to expand the dataset. We hope that these practical and methodological insights provide valuable guidance in extending the applicability and effectiveness of language model distillation in real-world scenarios.

\section*{Impact statement}

This paper presents work on language model distillation, which is actively used in the training of many modern language models. We identify a possible shortcoming of existing distillation procedures, called teacher hacking, that can lead to the transfer of unsafe behaviors from teacher to student. Additionally, we proposed several strategies to reduce the effect of this phenomenon. We believe that understanding and identifying such issues have positive societal consequences and allow the development of more reliable and safe language models.

\bibliography{reference}
\bibliographystyle{icml2025}

\newpage
\appendix
\onecolumn

\begin{figure}[ht]
    \centering
    \includegraphics[width=0.99\linewidth]{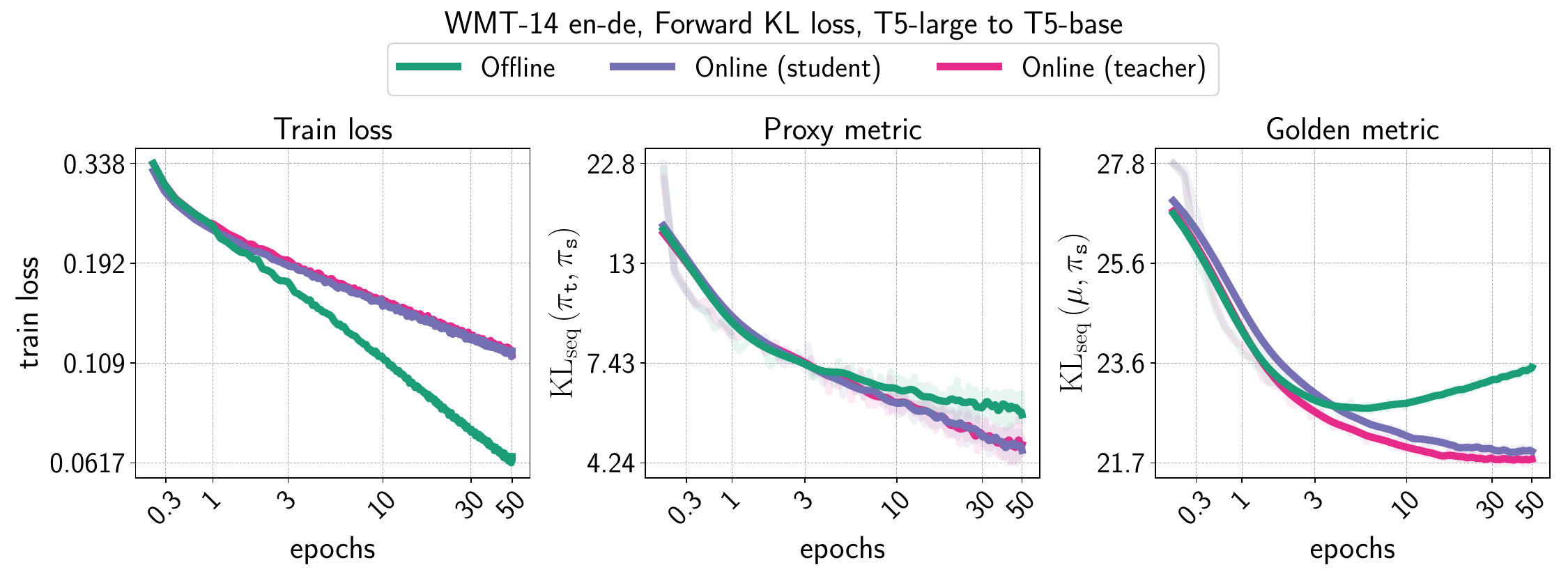}
    \includegraphics[width=0.99\linewidth]{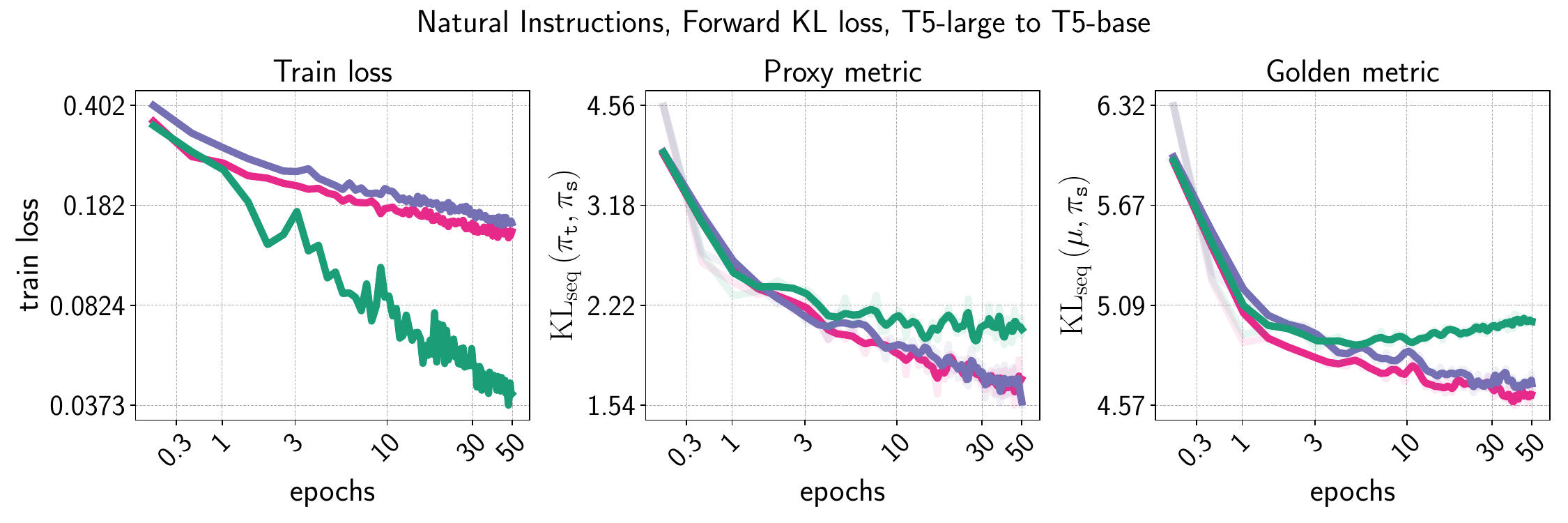}
    \caption{\textbf{Impact of the dataset choice: offline vs. online data sources.} 
    We verify our claims on the presence of teacher hacking in the case of offline data sources for two different tasks: the translation task on WMT-14 en-de (top row) and the instruction following task on Natural Instruction (bottom row). In general, the behavior of the curves is the same across all the datasets: for online data sources, both proxy and golden metrics are decreasing. At the same time, for offline data sources, the proxy metric is decreasing or stagnating, whereas the golden metric is clearly increasing.
    }
    \label{fig:datasets_exp_large_to_base_online_fwd_kl}
\end{figure}

\section{Additional experiments}\label{app:add_plots}

This section presents additional experiments on teacher hacking across various datasets, model sizes, and loss types, as well as an additional experiment on the effect of mixing offline and online data.

\begin{figure}[ht]
    \centering
    \includegraphics[width=0.99\linewidth]{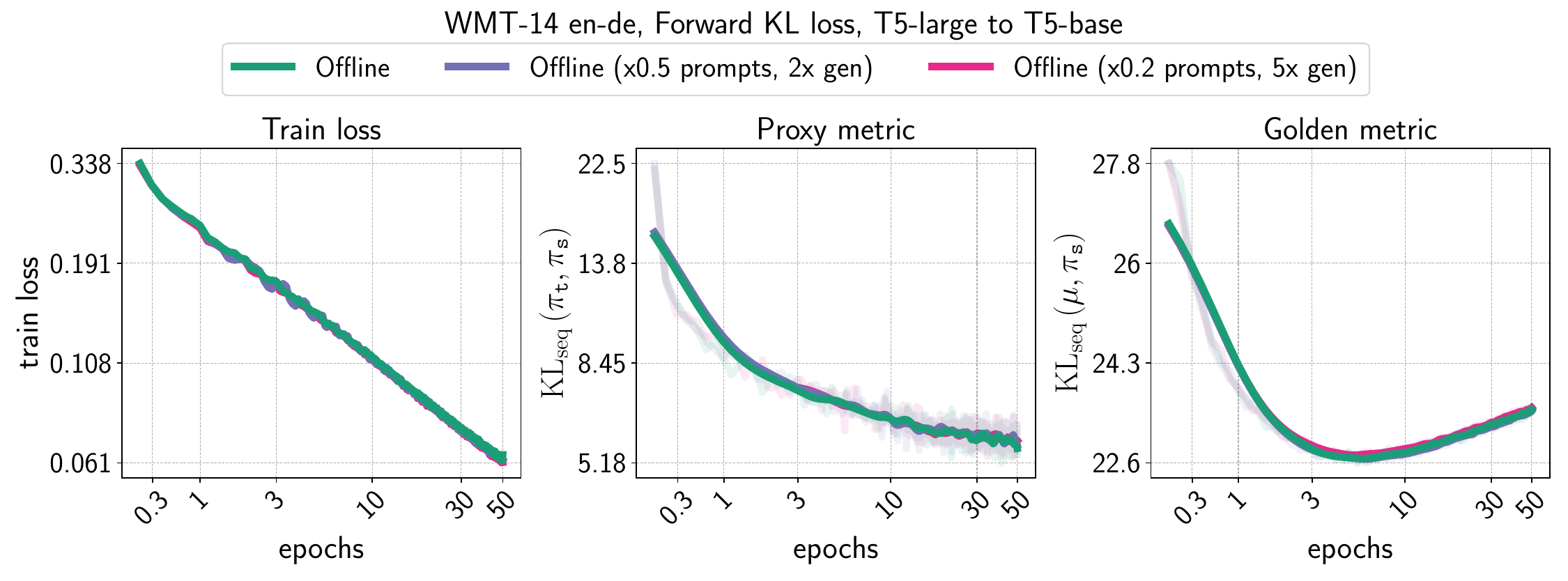}
    \includegraphics[width=0.99\linewidth]{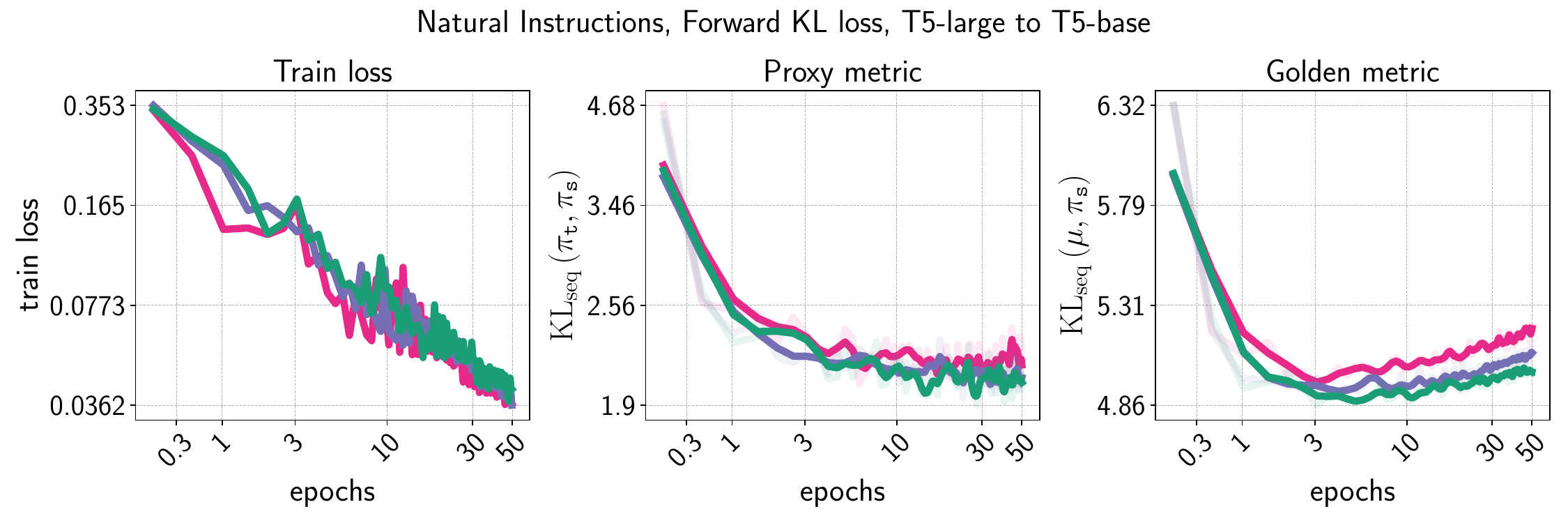}
    \caption{\textbf{Impact of the dataset choice: dataset diversity.} Across other datasets, we can notice that the impact of diversity is still present for the instruction-following task but not for the translation task. It can be explained by the initially small diversity of the WMT-14 dataset.
    }
    \label{fig:datasets_exp_large_to_base_diversity_fwd_kl}
\end{figure}

\subsection{The impact of the dataset}

In this subsection, we validate our claims across various tasks, such as the translation task on the WMT-14 en-de dataset \cite{bojar2014findings} and the instruction-following task on the Natural Instructions dataset \cite{naturalinstructions,supernaturalinstructions}. We consider the same pair of models: T5-large as the teacher model and T5-base as the student model, utilizing forward KL token-level loss along with the corresponding proxy and golden metrics.

\paragraph{Online vs. offline data sources.}
The results of the comparison of online and offline data sources on the WMT-14 en-de and Natural Instruction datasets are presented in \Cref{fig:datasets_exp_large_to_base_diversity_fwd_kl}. 

Overall, we observe the same phenomenon as noted in the summarization task in \Cref{sec:results}. The general patterns are consistent: for all data sources, the training loss decreases slowly for online data sources, as expected, while the proxy metric decreases or stagnates, showing no signs of classical overfitting. However, the golden metric continues to decline in the case of offline data sources, and it starts declining in the case of offline data source, indicating the presence of teacher hacking.

\paragraph{Diversity of offline data sources.}

Next, we study the impact of the diversity of the dataset on teacher hacking, following the setup described in \Cref{sec:diversity}. The results are presented in \Cref{fig:datasets_exp_large_to_base_diversity_fwd_kl}. We observe the same detrimental effect of the decreasing of the dataset diversity in the case of the instruction following task. However, in the case of the translation task, it almost has no effect. We could connect this effect to the small diversity or difficulty of the initial dataset since it contains only relatively short sentences.

\begin{figure}[ht]
    \centering
    \includegraphics[width=0.99\linewidth]{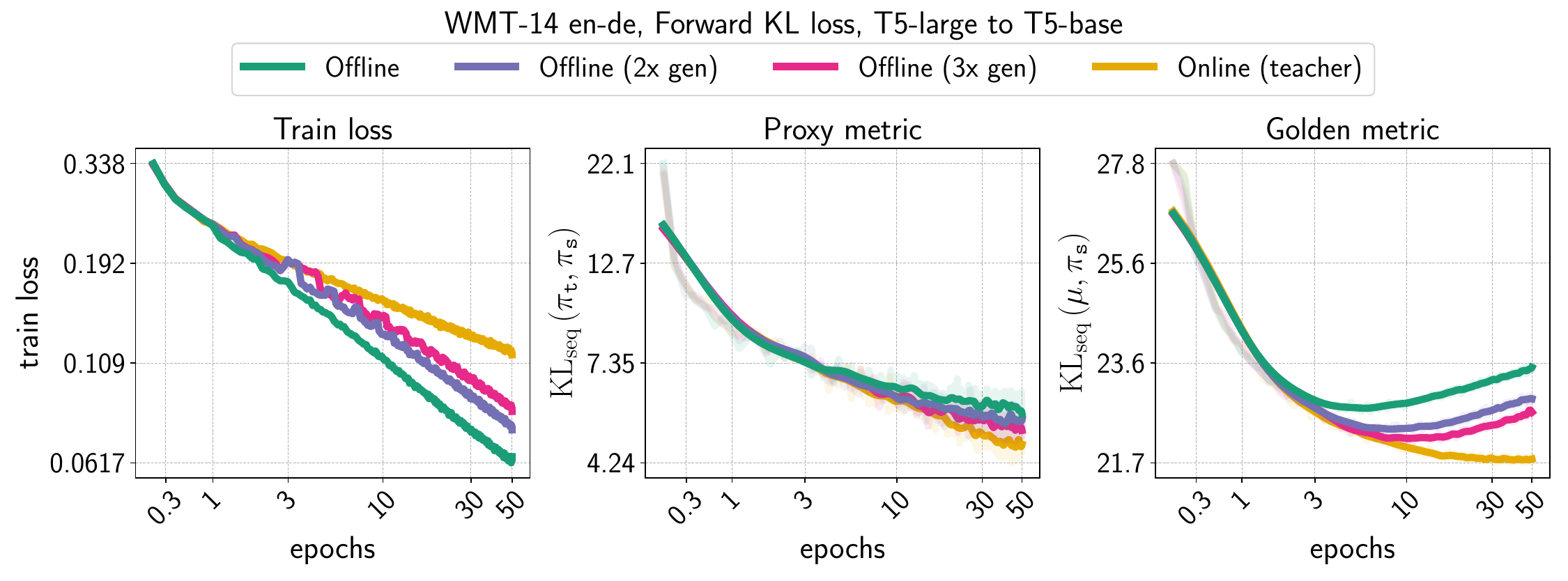}
    \includegraphics[width=0.99\linewidth]{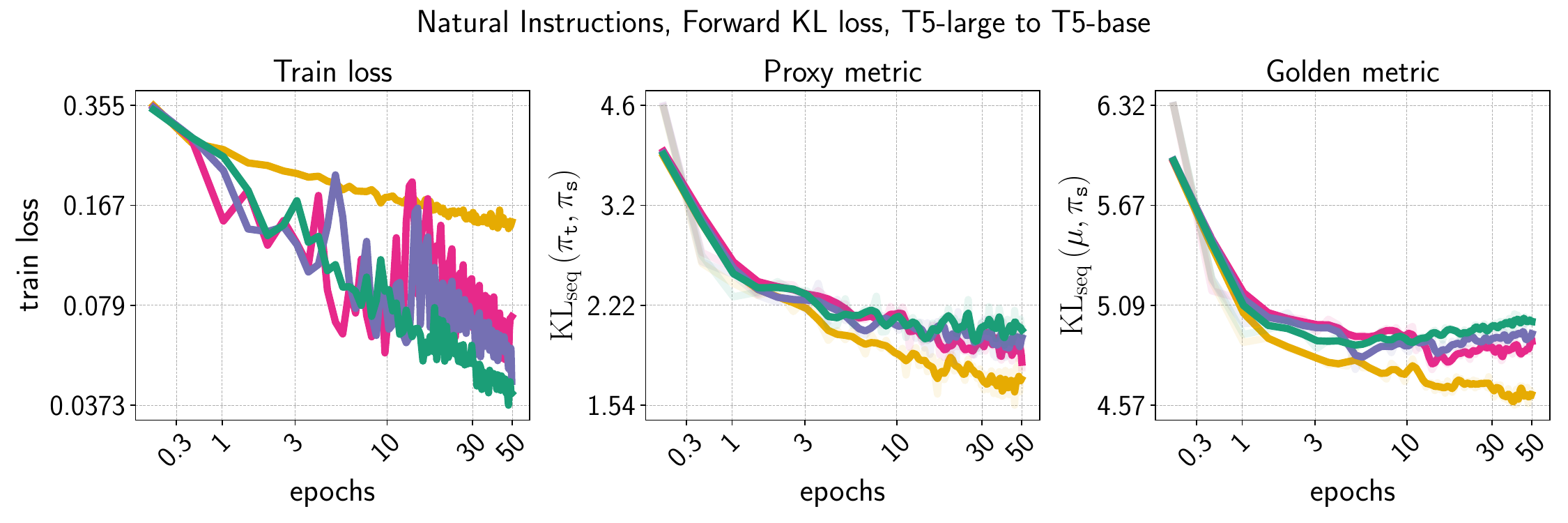}
    \caption{\textbf{Impact of the dataset choice: generation budget.} 
    We additionally confirm the claim on the positive impact of a larger number of generations per prompt across two other datasets.
    }
    \label{fig:datasets_exp_large_to_base_gen_fwd_kl}
\end{figure}

\paragraph{Generation budget.} Finally, we verify the claims on increasing the generation budget for the offline data sources. The results are presented in \Cref{fig:datasets_exp_large_to_base_gen_fwd_kl}. In this case, for both tasks, we observe an improvement in the golden metric, especially for the translation task, and we observe a marginal improvement in the proxy metric. As in the case of the summarization task, it signals the decreasing teacher hacking effect.

\subsection{The impact of student and teacher model sizes}

In this subsection, we validate our findings across different student and teacher model sizes using the XSum dataset and forward KL token-level loss.

\begin{figure}[ht]
    \centering
    \includegraphics[width=0.99\linewidth]{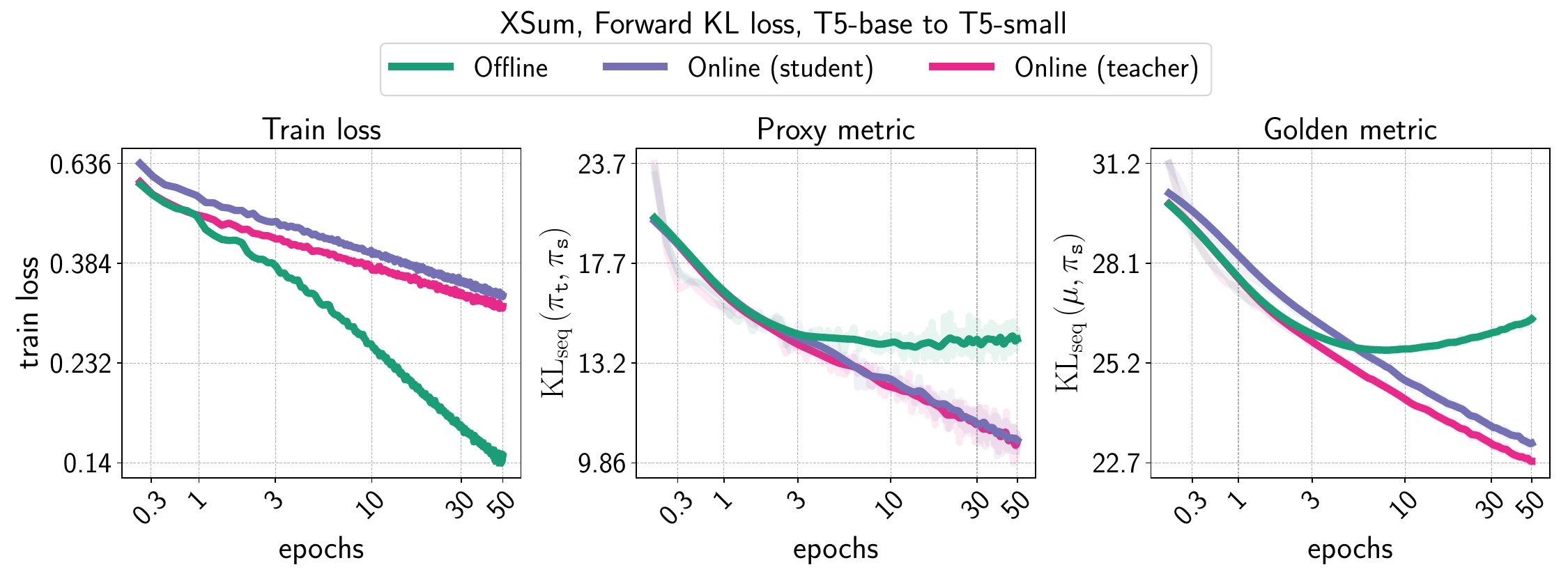}
    \includegraphics[width=0.99\linewidth]{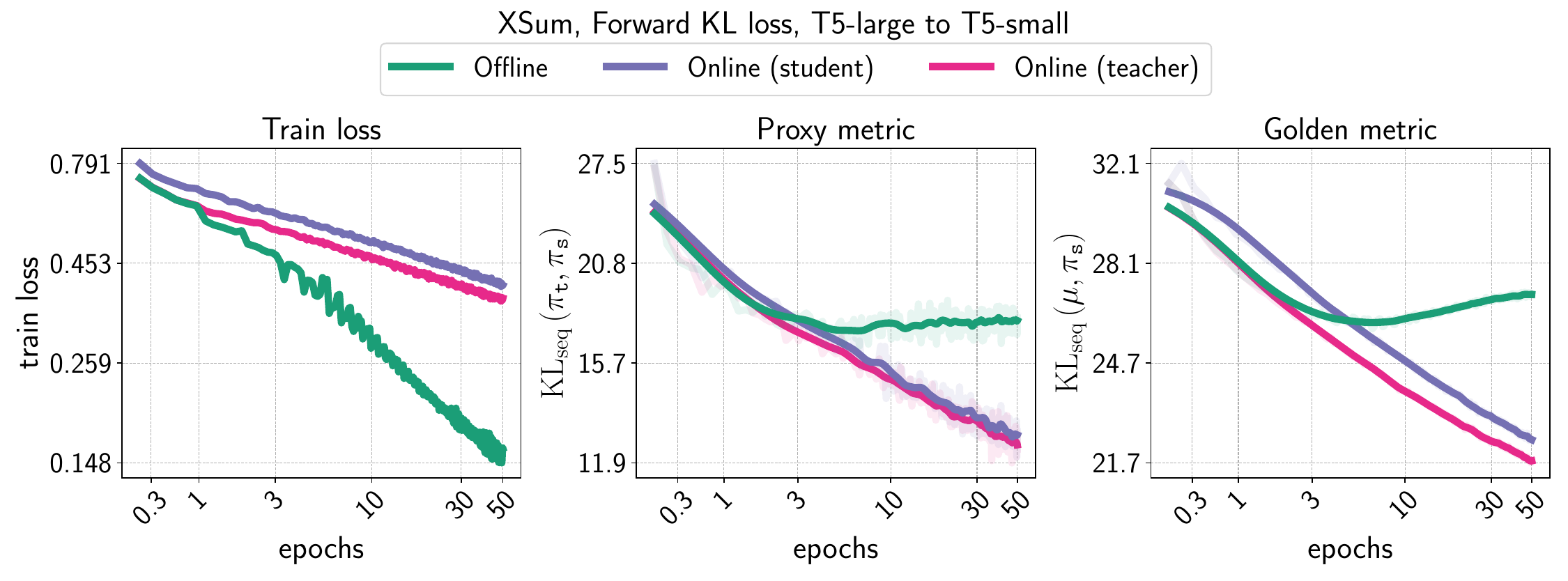}
    \caption{\textbf{Impact of the model sizes: offline vs. online data sources.} 
    We examine the train loss and proxy/golden metrics for two model size pairs: T5-base as the teacher and T5-small as the student (top), and T5-large as the teacher and T5-small as the student (bottom). For both pairs, no teacher hacking occurs with online data generation. However, teacher hacking is observed during distillation from T5-base to T5-small, as the proxy metric stagnates while the golden metric decreases. In contrast, distillation from T5-large to T5-small shows behavior consistent with standard overfitting as the proxy metric slightly increases. This may be attributed to the larger difference in model sizes.
    }
    \label{fig:sizes_exp_large_to_base_online_fwd_kl}
\end{figure}

\paragraph{Online vs. offline data sources.}
We conduct distillation experiments from T5-base to T5-small and from T5-large to T5-small, evaluating performance across offline and online data sources. The results are shown in \Cref{fig:sizes_exp_large_to_base_online_fwd_kl}.

For online data sources, both the proxy and golden metrics decrease monotonically, regardless of the model sizes, confirming our earlier observations. In contrast, for offline data sources, the golden metric consistently increases. Notably, in the case of distilling T5-large to T5-small, the proxy metric also shows a slight increase, indicating standard overfitting rather than teacher hacking. Meanwhile, when distilling T5-base to T5-small, the proxy metric stagnates, suggesting the presence of teacher hacking.

\subsection{The impact of token-level loss functions}

\begin{figure}[ht]
    \centering
    \includegraphics[width=0.99\linewidth]{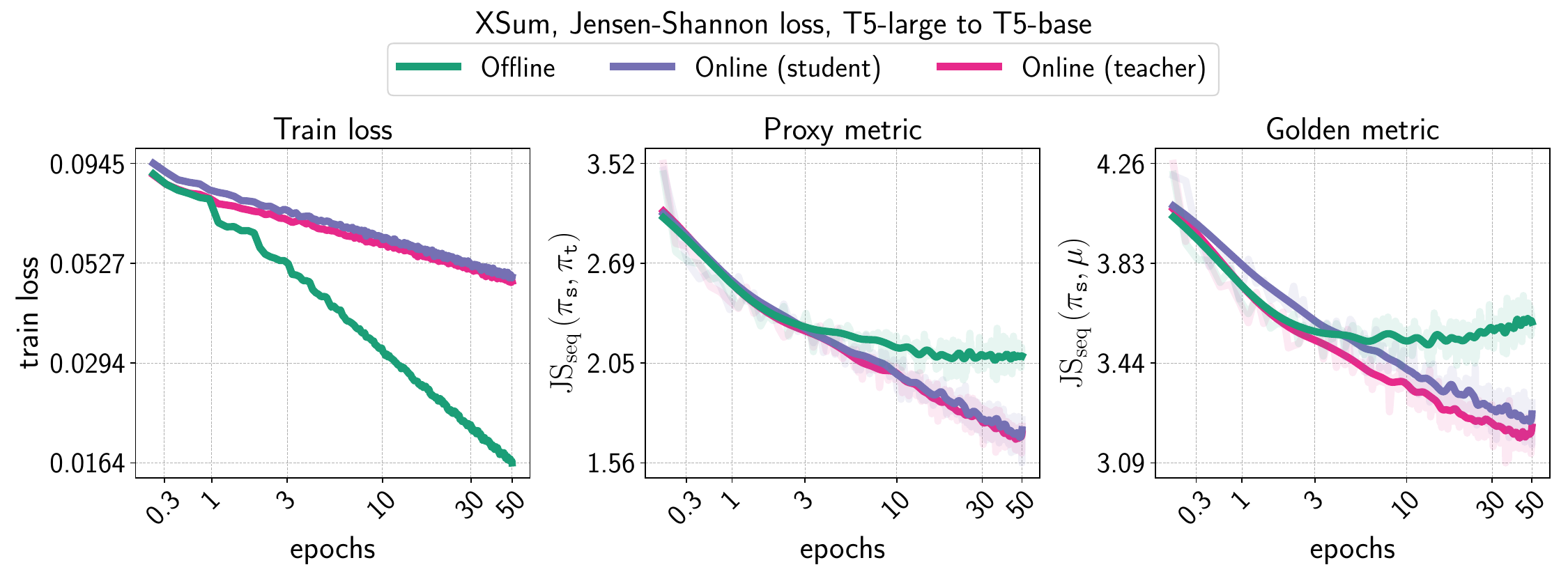}
    \includegraphics[width=0.99\linewidth]{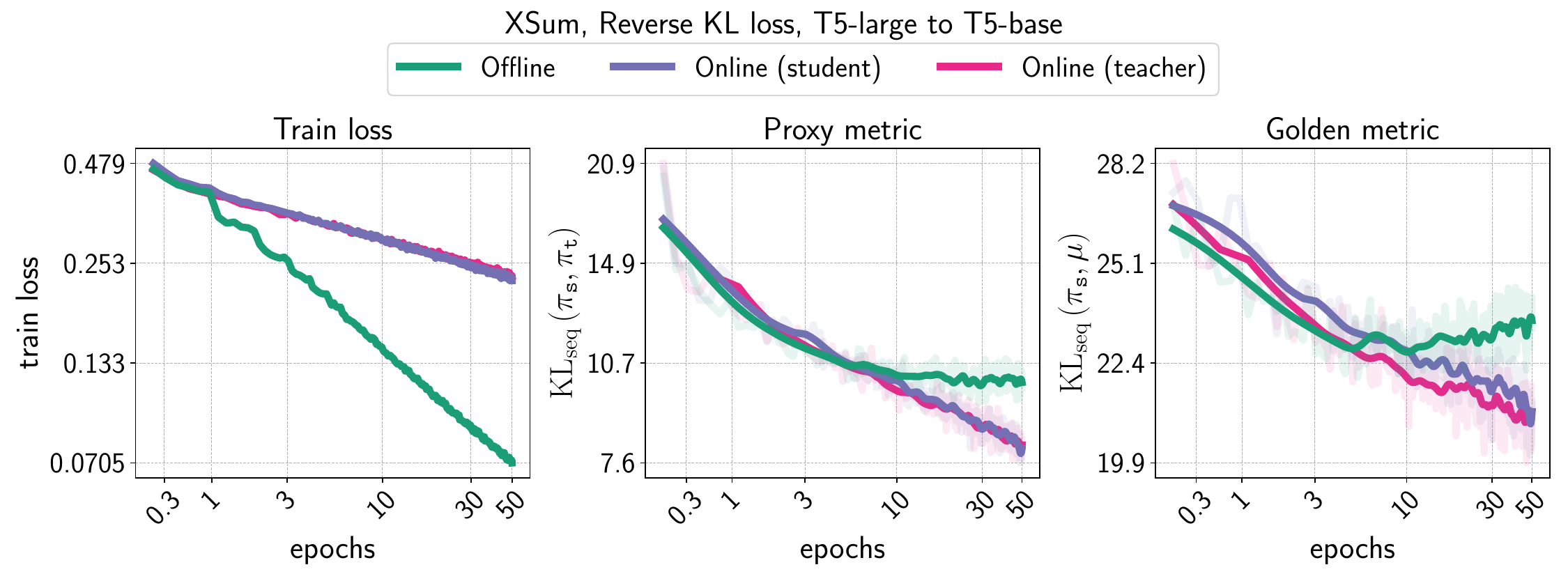}
    \caption{\textbf{Impact of the loss type: offline vs. online data sources.} 
    We can observe that the effect of teacher hacking appears regardless of the choice of loss function.
    }
    \label{fig:losses_exp_large_to_base_online_fwd_kl}
\end{figure}

In this subsection, we examine how the choice of token-level loss function and proxy/golden metrics influences the teacher hacking phenomenon. The experiments are conducted using the XSum summarization dataset, with T5-base as the student model and T5-large as the teacher model.

\paragraph{Online vs. offline data sources.}
This experiment compares different data sources using two token-level loss functions: Jensen-Shannon divergence and reverse KL divergence, alongside their corresponding proxy and golden metrics. The results are shown in \Cref{fig:losses_exp_large_to_base_online_fwd_kl}.

The observed behavior across all plots closely resembles that of the forward KL loss. Specifically, when using online data sources, both proxy and golden metrics decrease monotonically. In contrast, for offline data sources, proxy metrics continue to decrease, but the golden metric increases, signaling the presence of teacher hacking.

Additional experiments were conducted using generalized Jensen-Shannon divergences with coefficients of $0.1$ and $0.9$. However, their performance was either comparable to or worse than reverse or forward KL divergence for the respective proxy and golden metrics. As a result, they are excluded from the comparison.

\subsection{Offline-online data mixtures}

In this subsection, we examine how varying mixtures of offline and online data influence the occurrence of teacher hacking. The experiment uses the XSum dataset, T5-base as the student model, T5-large as the teacher model, and forward KL token-level loss, following \Cref{sec:results}.

\begin{figure}[ht]
    \centering
    \includegraphics[width=0.99\linewidth]{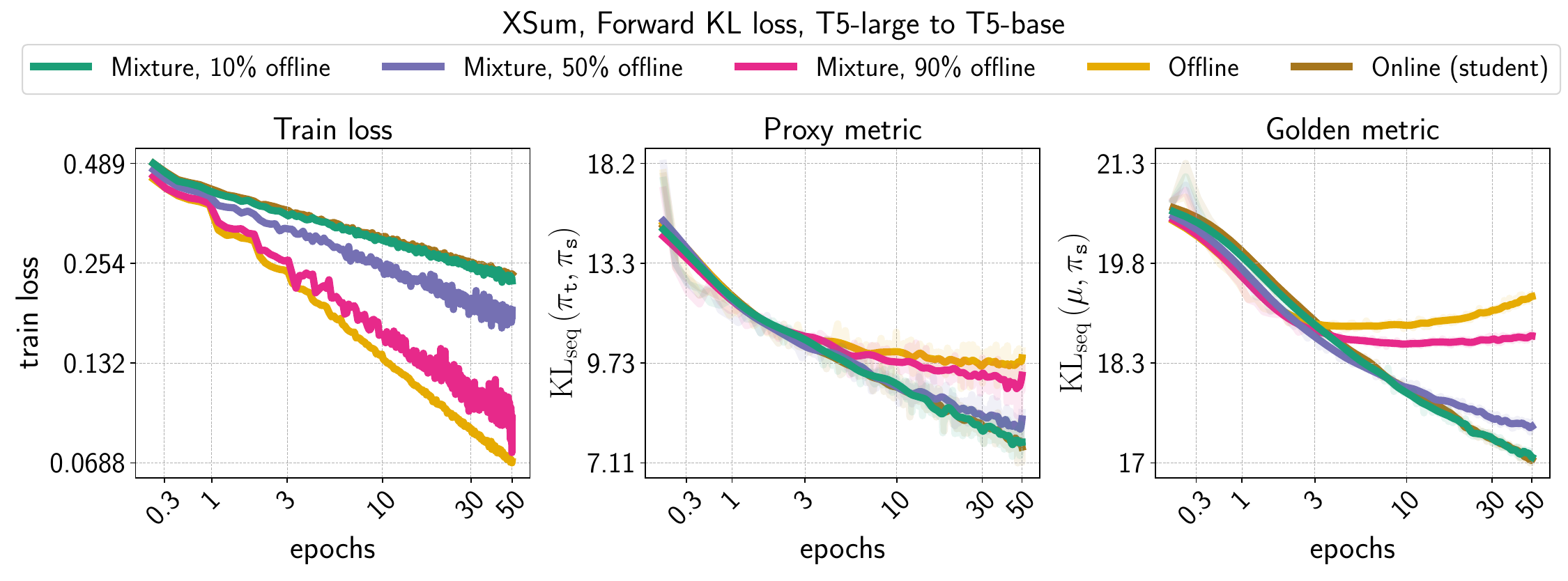}
    \caption{\textbf{Mixture of offline and online data.} 
    This plot compares strategies for combining offline and online data during the distillation process. The results show that incorporating just 10\% online student data significantly reduces the effect of teacher hacking, causing the golden metric to stabilize rather than increase.  At the same time, the usage of at least 50\% of the online generated data allows to avoid the effect of teacher hacking completely.
    }
    \label{fig:mixture_exp_large_to_base_online_fwd_kl}
\end{figure}

We use the following procedure to generate each training batch: with probability $\alpha$, the batch is sampled from a fixed offline dataset, and with probability $1-\alpha$, it is generated by the current student model. This process is repeated at each training step. We evaluate three values for $\alpha$: 0.1, 0.5, and 0.9, corresponding to 10\%, 50\%, and 90\%  offline data proportion, respectively. The results are shown in \Cref{fig:mixture_exp_large_to_base_online_fwd_kl}.

Our results show that increasing the proportion of \textit{online} data in the distillation process (that is equivalent to decreasing the proportion of offline data) significantly improves the golden metric. Even with 10\% online student data (corresponding to 90\% offline data), the golden metric plateaus, thus effectively reducing teacher hacking. Higher proportions of online data further mitigate the effect, with 90\% online student data resulting in training dynamics nearly identical to those observed when using only student-generated data.

\paragraph{Discussion.} These results suggest that exclusively using online-generated data is not required to avoid teacher hacking. Instead, incorporating a fraction of online-generated data during the distillation process is sufficient. In particular, as little as 10\% online data can substantially reduce the impact of teacher hacking.

\section{Additional details}\label{app:misc}

In this section, we provide an exact formulation for sequence-level divergences
\begin{equation}\label{eq:kl_divergence_lms}
    \KL_{\seq}(\pi,  \pi') \triangleq \E_{x \sim d(\cdot), y \sim p_{\pi}(\cdot | x)}\left[ \sum_{i=1}^{|y|}\KL(\pi(\cdot | x, y_{:i})), \pi'(\cdot | x, y_{:i})) \right]\,.
\end{equation}
In particular, to estimate this KL-divergence, we need to sample prompts $x$ and generate responses using a language model $\pi$. Additionally, we introduce a sequence-based Jensen-Shannon divergence, defined as
\begin{equation}\label{eq:js_divergence_lms}
    \begin{split}
    \JS_{\seq}(\pi, \pi') &\triangleq \E_{x \sim d(\cdot), y \sim p_{\pi}(\cdot | x), y' \sim p_{\pi'}(\cdot | x)}\bigg[ \frac{1}{2}\sum_{i=1}^{|y|}\KL(\pi(\cdot | x, y_{:i}),  m(\cdot | x, y_{:i}))  + \frac{1}{2}\sum_{i=1}^{|y'|}\KL(\pi'(\cdot | x, y'_{:i}), m(\cdot | x, y'_{:i})) \bigg]\,,
    \end{split}
\end{equation}
where $m(\omega | x, y) \triangleq 0.5 \cdot \pi(\omega| x,y) + 0.5 \cdot \pi'(\omega|x,y)$ is a mixture of two language models. To estimate this divergence, we need to use samples from both models $\pi$ and $\pi'$ and compute an average of two token-level KL-divergences. Notice that computation of a true Jensen-Shannon divergence between $p_{\pi}$ and $p_{\pi'}$ is computationally infeasible since the log-probabilities of the mixture $0.5 \cdot p_{\pi}(y|x) + 0.5 \cdot p_{\pi'}(y|x)$ does not satisfy a chain rule in terms of $\pi$ and $\pi'$.
\section{Hyperparameters}\label{sec:hyperparameters}

In this section, we provide detailed information on the hyperparameters used for our experiments; see \Cref{tab:hyperparams_xsum_wmt} and \Cref{tab:hyperparams_natural_instructions}. 

\begin{table}[h!]
\centering
\caption{Hyperparameter details for summarization \& translation tasks.}
\begin{tabular}{@{}ll@{}}
\toprule
\textbf{Hyperparameter} & \textbf{Value} \\ \midrule
Oracle Dataset Size     & 100,000       \\
Distillation Dataset Size & 200,000     \\
Training Steps          & 390,625         \\
Batch Size              & 32            \\
Task                    & XSum \\
Dropout                 & 0.0            \\
Warmup Schedule         & 100 steps      \\
Optimal Learning Rate (LR)  & 0.0003         \\
Input Length (Tokenized)    & 1024           \\
Output Length (Tokenized)   & 128           \\
Softmax Temperature     & 1.0      \\ \bottomrule
\end{tabular}
\qquad\qquad\qquad
\begin{tabular}{@{}ll@{}}
\toprule
\textbf{Hyperparameter} & \textbf{Value} \\ \midrule
Oracle Dataset Size     & 50,000       \\
Distillation Dataset Size & 450,000     \\
Training Steps          & 703,125         \\
Batch Size              & 32            \\
Task                    & WMT-14 en-de  \\
Dropout                 & 0.0            \\
Warmup Schedule         & 100 steps      \\
Optimal Learning Rate (LR)  & 0.0003         \\
Input Length (Tokenized)    & 80           \\
Output Length (Tokenized)   & 80           \\
Softmax Temperature     & 1.0      \\ \bottomrule
\end{tabular}
\label{tab:hyperparams_xsum_wmt}
\end{table}

\begin{table}[h!]
\centering
\caption{Hyperparameter details for instruction following task.}
\begin{tabular}{@{}ll@{}}
\toprule
\textbf{Hyperparameter} & \textbf{Value} \\ \midrule
Oracle Dataset Size     & 100,000       \\
Distillation Dataset Size & 500,000     \\
Training Steps          & 390,625         \\
Batch Size              & 64            \\
Task                    & Natural Instructions  \\
Dropout                 & 0.0            \\
Warmup Schedule         & 100 steps      \\
Optimal Learning Rate (LR)  & 0.0003         \\
Input Length (Tokenized)    & 2048           \\
Output Length (Tokenized)   & 256           \\
Softmax Temperature     & 1.0      \\ \bottomrule
\end{tabular}
\label{tab:hyperparams_natural_instructions}
\end{table}


\end{document}